\pdfoutput=1

\documentclass[11pt]{article}

\usepackage[]{acl}

\usepackage{times}
\usepackage{latexsym}

\usepackage[T1]{fontenc}

\usepackage[utf8]{inputenc}

\usepackage{microtype}

\usepackage{graphicx}

\title{BLAB Reporter: Automated journalism covering the Blue Amazon}

\author{Yan Vianna Sym \\
Escola Politécnica \\ Universidade de São Paulo \\ São Paulo, Brazil \\
\texttt{yan.sym@usp.br} \\\And
João Gabriel Moura Campos \\
Escola Politécnica \\ Universidade de São Paulo \\ São Paulo, Brazil \\
\texttt{joaogcampos@usp.br} \\\And
Fabio Gagliardi Cozman \\
Escola Politécnica \\ Universidade de São Paulo \\ São Paulo, Brazil \\
\texttt{fgcozman@usp.br}}

\begin{document}

\maketitle
\begin{abstract}
This demo paper introduces the BLAB Reporter, a robot-journalist covering the Brazilian Blue Amazon. The Reporter is based on a pipeline architecture for Natural Language Generation; it offers daily reports, news summaries and curious facts in Brazilian Portuguese. By collecting, storing and analysing structured data from publicly available sources, the robot-journalist uses domain knowledge to generate  and publish texts in Twitter. Code and corpus are publicly available \footnote{\url{https://github.com/C4AI/blab-reporter}}.
\end{abstract}

\section{Introduction}

Data-to-text Natural Language Generation (NLG) is the computational process of generating meaningful and coherent natural text or speech to describe non-linguistic input data \cite{Ehud}. Successful examples of data-to-text systems can be found in both academia and industry, with applications in weather forecasting \cite{belz2008automatic}, image captions and chatbots \cite{adamopoulou2020overview}. Amongst NLG applications, robot-journalism is one of the most prominent endeavors thanks to the high volume of structured data streams available, which enables automated systems to report recurrent information with high-fidelity and lexical variety \cite{teixeira2020damata}.

An interesting domain for data-to-text generation is ocean monitoring. For instance, global attention was drawn in 2021 to a container ship that obstructed the Suez Canal for six consecutive days, causing a global shortage of essential commodities, including medical supplies and medicines during the coronavirus pandemic. Accurate and low latency information reports can be very helpful in these situations, but communicating to general audiences  usually demands coverage by specialized human journalists. To address this issue, we present our robot-journalist named \textit{BLAB Reporter}, a NLG system based on a pipeline architecture that generates daily reports, news, content summarization and curious facts about the Blue Amazon and publishes them on Twitter in Brazilian Portuguese \footnote{\url{https://twitter.com/BLAB_Reporter}}. The Blue Amazon is the exclusive economic zone (EEZ) of Brazil, with an offshore area of 3.6 million square kilometers along the Brazilian coast, an area rich in marine biodiversity and energy resources \cite{wiesebron2013blue}. The BLue Amazon Brain (BLAB) is a project aiming to address complex questions about the marine ecosystem, and integrates a number of services aimed at disseminating information about the Blue Amazon.

\section{System overview}

    \begin{figure*}[ht]
        \centering
        \includegraphics[width=0.97\textwidth]{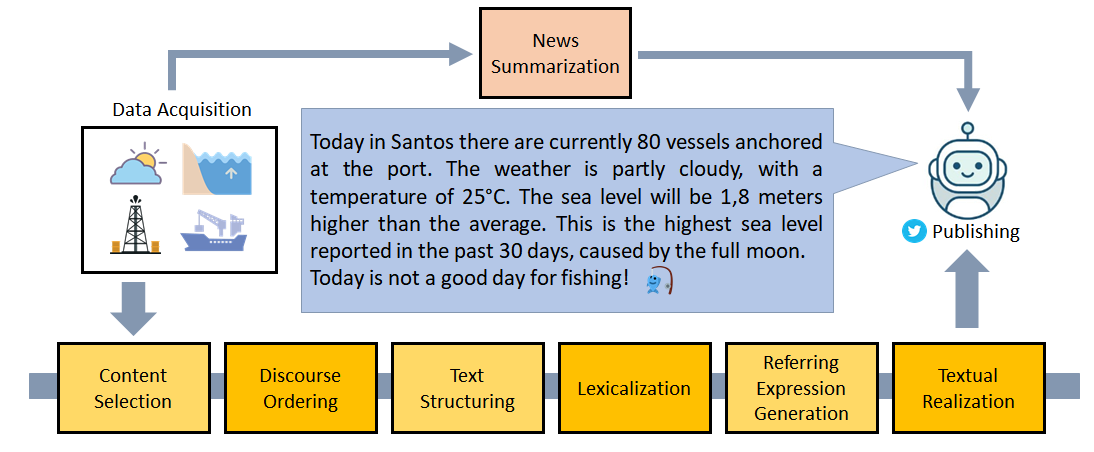}
        \vspace*{-2ex}
        \caption{Robot-journalist system architecture}
        \label{figura1}
    \end{figure*}

Our system follows a pipeline architecture that converts non-linguistic data into text in 6 steps: \textit{Content Selection, Discourse Ordering, Text Structuring, Lexicalization, Referring Expression Generation and Textual Realization} \cite{ferreira2019neural}. Our system also comprises two additional steps: \textit{Data Acquisition} (for extracting and storing information from multiple data streams in a structured format) and \textit{Summarization} (for summarizing news in the form of small consecutive tweets). This kind of architecture, depicted in Figure \ref{figura1}, allows for trustworthy output as well as easy access to and maintenance of sub-modules.

The grammar used by the model was built by first running the content selection step in previous data and generating 30 non-linguistic reports. These non-linguistic reports were then manually verbalized and the input and output representations for each pipeline module were manually annotated. When deployed, each module draws on the selected combination of templates using rule-based approaches. Because we deal with a sensitive domain, we opted to use the pipeline architecture instead of the novel end-to-end systems, which sometimes hallucinates content \cite{ji2022survey}. The following sections describe each module.

\textbf{Data Acquisition}  The first step of our system performs data gathering, filtering and cleaning before it is put in a data warehouse. In our application, this module consists of a web scraping framework for extracting information from public websites and storing it on a structured format. Our system currently collects information about weather, tide charts, marine vessel traffic and eventual earthquakes on the Brazilian coast, and stores data using MongoDB, a source-available cross-platform document-oriented database program \cite{gyHorodi2015comparative}.

\textbf{Content Selection}  This module decides which relevant information should be verbalized in the text. The content selection process generally consists of applying domain specific knowledge to create a rule-based approach. The following text is an example of the content selection module output: 
    {\tt \footnotesize \\ CURRENT WEATHER AND TEMPERATURE (weather="partly cloudy",temperature="25ºC",city="Santos", timestamp="May 22, 2022); FISHING CONDITION (condition="good",event="sea level is high";height of the sea:"1.8 meters";days since last peak="30");\\ CAUSE(earthquake:"no",moon calendar:"yes");}

\textbf{Discourse Ordering} and \textbf{Text Structuring}  Once the relevant content has been selected, our application constructs a logical timeline of events and sorts the intent messages in sentences and paragraphs in order to enhance reader comprehension \cite{heilbron2019tracking}. This combined module is based on a list of possible intent orderings collected from the corpus and decides what is the most optimal way to sort the sentences, bearing in mind the 280 character limit on Twitter. For example, for the messages related to weather conditions, a possible outcome order would be:

\noindent
{\small {\tt WEATHER ALERT → CAUSES → DAYS SINCE LAST PEAK}}

\textbf{Lexicalization}  At this step of the pipeline, lexical choices are made in order to verbalize the intents, finding the proper words to generate proper sentences. We applied a template-based lexicalization with plenty of options to choose from, providing for more inflections and variety of text in comparison to the fill-template approach \cite{stede1994lexicalization}. Templates provide for gender and number inflection, for example: \textit{“No Rio de Janeiro foi registrada a maior temperatura da última semana”} vs. \textit{"Em São Paulo foi registrado o maior vento dos últimos 10 dias"}.

\textbf{Referring Expression Generation}  In order to replace entity tags throughout the template, this module generates the appropriate references using a list of possible expressions for each entity \cite{krahmer2012computational}. For the first reference to an entity in the text, a full description is used
(e.g., INSTITUTE → "The Seismological Center at the University of São Paulo (USP)"), whereas for subsequent references a random referring expression
to the entity is chosen (e.g., INSTITUTE → "The Seismological Center at USP"; "the Institute"; "sismoUSP"; "it"; etc.).

\textbf{Textual Realization}  The last step of the pipeline is responsible for transforming intermediate representations into human-readable Brazilian Portuguese text. After the content is generated, this step applies a final rule based transformation to the text with the goal to make the texts look more natural, for example adding greetings message and emojis. We also added a validation layer in this step, to ensure there is no offensive content within the text. The output of this module is published using Twitter's API. An example of generated text is shown in Figure \ref{figura1}.

\textbf{Summarization}  An extra module was implemented in our system in order to outline public news about the Blue Amazon while also splitting text into small consecutive tweets. Because data hallucination is less critical in this step, this module was implemented using PTT5, a T5 model pretrained in a large collection of web pages in Portuguese, which uses state of the art transformer architecture \cite{carmo2020ptt5}. Key challenges of this approach are interpretation and evaluation of the generated texts \cite{rao2018computational}.

The generated texts are  published on specific periods of time. We have noticed that weather related content has more user engagement during mornings, while news and curious facts content are usually more viewed in the evenings. More critical messages, for example information related to earthquakes in the Blue Amazon region, are published as soon as the data is collected.

\section{Conclusions}

This paper presents a data-to-text system based on a pipeline architecture for NLG. Our system applies robot-journalism techniques to generate and publish reports, news and curious facts in Brazilian Portuguese about the Blue Amazon. Due to its rule-base nature, our system provides high-fidelity content by applying a pipeline methodology and obtains lexical variety by drawing from a list of multiple available template options for the same intent. In the future we plan to add more sources of information to the pipeline, for example statistics about oil exploration and reporting of illegal fishing activities in real time. We also plan to utilize user engagement data and apply artificial neural network techniques to improve our system's performance.

\subsection*{Acknowledgements}

We would like to thank the Center for Artificial Intelligence (C4AI: www.c4ai.inova.usp) with support from the São Paulo Research Foundation (FAPESP grant 2019/07665-4) and from IBM Corporation. The third author was partially supported by CNPq grant 312180/2018-7 (PQ).

\bibliography{anthology,custom}

\begin{thebibliography}{13}
\expandafter\ifx\csname natexlab\endcsname\relax\def\natexlab#1{#1}\fi

\bibitem[{Adamopoulou and Moussiades(2020)}]{adamopoulou2020overview}
Eleni Adamopoulou and Lefteris Moussiades. 2020.
\newblock An overview of chatbot technology.
\newblock In \emph{IFIP International Conference on Artificial Intelligence
  Applications and Innovations}, pages 373--383. Springer.

\bibitem[{Belz(2008)}]{belz2008automatic}
Anja Belz. 2008.
\newblock Automatic generation of weather forecast texts using comprehensive
  probabilistic generation-space models.
\newblock \emph{Natural Language Engineering}, 14(4):431--455.

\bibitem[{Carmo et~al.(2020)Carmo, Piau, Campiotti, Nogueira, and
  Lotufo}]{carmo2020ptt5}
Diedre Carmo, Marcos Piau, Israel Campiotti, Rodrigo Nogueira, and Roberto
  Lotufo. 2020.
\newblock Ptt5: Pretraining and validating the t5 model on brazilian portuguese
  data.
\newblock \emph{arXiv preprint arXiv:2008.09144}.

\bibitem[{Ferreira et~al.(2019)Ferreira, van~der Lee, Van~Miltenburg, and
  Krahmer}]{ferreira2019neural}
Thiago~Castro Ferreira, Chris van~der Lee, Emiel Van~Miltenburg, and Emiel
  Krahmer. 2019.
\newblock Neural data-to-text generation: A comparison between pipeline and
  end-to-end architectures.
\newblock \emph{arXiv preprint arXiv:1908.09022}.

\bibitem[{Gy{\H{o}}r{\"o}di et~al.(2015)Gy{\H{o}}r{\"o}di, Gy{\H{o}}r{\"o}di,
  Pecherle, and Olah}]{gyHorodi2015comparative}
Cornelia Gy{\H{o}}r{\"o}di, Robert Gy{\H{o}}r{\"o}di, George Pecherle, and
  Andrada Olah. 2015.
\newblock A comparative study: Mongodb vs. mysql.
\newblock In \emph{2015 13th International Conference on Engineering of Modern
  Electric Systems (EMES)}, pages 1--6. IEEE.

\bibitem[{Heilbron et~al.(2019)Heilbron, Ehinger, Hagoort, and
  De~Lange}]{heilbron2019tracking}
Micha Heilbron, Benedikt Ehinger, Peter Hagoort, and Floris~P De~Lange. 2019.
\newblock Tracking naturalistic linguistic predictions with deep neural
  language models.
\newblock \emph{arXiv preprint arXiv:1909.04400}.

\bibitem[{Ji et~al.(2022)Ji, Lee, Frieske, Yu, Su, Xu, Ishii, Bang, Madotto,
  and Fung}]{ji2022survey}
Ziwei Ji, Nayeon Lee, Rita Frieske, Tiezheng Yu, Dan Su, Yan Xu, Etsuko Ishii,
  Yejin Bang, Andrea Madotto, and Pascale Fung. 2022.
\newblock Survey of hallucination in natural language generation.
\newblock \emph{arXiv preprint arXiv:2202.03629}.

\bibitem[{Krahmer and Van~Deemter(2012)}]{krahmer2012computational}
Emiel Krahmer and Kees Van~Deemter. 2012.
\newblock Computational generation of referring expressions: A survey.
\newblock \emph{Computational Linguistics}, 38(1):173--218.

\bibitem[{Rao and Gudivada(2018)}]{rao2018computational}
CR~Rao and Venkat~N Gudivada. 2018.
\newblock \emph{Computational analysis and understanding of natural languages:
  principles, methods and applications}.
\newblock Elsevier.

\bibitem[{Reiter and Dale(2000)}]{Ehud}
Ehud Reiter and Robert Dale. 2000.
\newblock Building applied natural language generation systems.
\newblock \emph{Natural Language Engineering}.

\bibitem[{Stede(1994)}]{stede1994lexicalization}
Manfred Stede. 1994.
\newblock Lexicalization in natural language generation: A survey.
\newblock \emph{Artificial Intelligence Review}, 8(4):309--336.

\bibitem[{Teixeira et~al.(2020)Teixeira, Campos, Cunha, Ferreira, Pagano, and
  Cozman}]{teixeira2020damata}
Andr{\'e} Luiz~Rosa Teixeira, Jo{\~a}o Campos, Rossana Cunha, Thiago~Castro
  Ferreira, Adriana Pagano, and Fabio Cozman. 2020.
\newblock {DaMata}: A robot-journalist covering the brazilian amazon
  deforestation.
\newblock In \emph{Proceedings of the 13th International Conference on Natural
  Language Generation}, pages 103--106.

\bibitem[{Wiesebron(2013)}]{wiesebron2013blue}
Marianne Wiesebron. 2013.
\newblock Blue {A}mazon: thinking about the defence of the maritime territory.
\newblock \emph{Austral: Brazilian Journal of Strategy \& International
  Relations}, 2(3):107--132.

\end{thebibliography}
\bibliographystyle{acl_natbib}

\end{document}